\documentclass[10pt,twocolumn,letterpaper]{article}

\usepackage{iccv}
\usepackage{times}
\usepackage{epsfig}
\usepackage{graphicx}
\usepackage{amsmath}
\usepackage{amssymb}
\usepackage{booktabs}
\usepackage{subcaption}
\usepackage{bm}
\usepackage{float}
\usepackage{slashbox}
\iccvfinalcopy
\def\iccvPaperID{12675} % Enter the WACV Paper ID here

\ificcvfinal\fi

\usepackage[pagebackref=true,breaklinks=true,letterpaper=true,colorlinks,bookmarks=false]{hyperref}

\usepackage[capitalize]{cleveref}
\crefname{section}{Sec.}{Secs.}
\Crefname{section}{Section}{Sections}
\Crefname{table}{Table}{Tables}
\crefname{table}{Tab.}{Tabs.}
\begin{document}

%%%%%%%%% TITLE
\title{ AutoDepthNet: High Frame Rate Depth Map Reconstruction\\ using Commodity Depth and RGB Cameras}
% \author{First Author\\
% Institution1\\
% Institution1 address\\
% {\tt\small firstauthor@i1.org}
\author{Peyman Gholami \qquad Robert Xiao\\
Department of Computer Science, University of British Columbia, Canada\\
{\tt\small \{peymang, brx\}@cs.ubc.ca}
% For a paper whose authors are all at the same institution,
% omit the following lines up until the closing ``}''.
% Additional authors and addresses can be added with ``\and'',
% just like the second author.
% To save space, use either the email address or home page, not both
% \and
% Robert Xiao\\
% University of British Columbia\\
% Department of Computer Science\\
% {\tt\small brx@cs.ubc.ca}
}
\maketitle
\thispagestyle{empty}

%%%%%%%%% ABSTRACT
\begin{abstract}
Depth cameras have found applications in diverse fields, such as computer vision, artificial intelligence, and video gaming. However, the high latency and low frame rate of existing commodity depth cameras impose limitations on their applications. We propose a fast and accurate depth map reconstruction technique to reduce latency and increase the frame rate in depth cameras. Our approach uses only a commodity depth camera and color camera in a hybrid camera setup; our prototype is implemented using a Kinect Azure depth camera at 30 fps and a high-speed RGB iPhone 11 Pro camera captured at 240 fps. The proposed network, AutoDepthNet, is an encoder-decoder model that captures frames from the high-speed RGB camera and combines them with previous depth frames to reconstruct a stream of high frame rate depth maps. On GPU, with a 480$\times$270 output resolution, our system achieves an inference time of 8 ms, enabling real-time use at up to 200 fps with parallel processing. AutoDepthNet can estimate depth values with an average RMS error of 0.076, a 44.5\% improvement compared to an optical flow-based comparison method. Our method can also improve depth map quality by estimating depth values for missing and invalidated pixels. The proposed method can be easily applied to existing depth cameras and facilitates the use of depth cameras in applications that require high-speed depth estimation. We also showcase the effectiveness of the framework in upsampling different sparse datasets \eg video object segmentation. As a demonstration of our method, we integrated our framework into existing body tracking systems and demonstrated the robustness of the proposed method in such applications.
\end{abstract}
%%%%%%%%% BODY TEXT
\section{Introduction}
\label{sec:intro}
Low-cost, commodity depth sensing cameras are a key advancement in 3D sensing technologies and have proven useful in many research areas, including robotics~\cite{du2012robotics}, augmented and virtual reality (AR/VR)~\cite{capece2018arvr}, object detection, tracking, and recognition~\cite{ni2013objectdetection}, autonomous driving~\cite{hernandez2016drive}, \etc. 

Depth cameras vary based on their sensing principle. Common types include stereo matching sensors, time-of-flight (ToF) sensors, and light detection and ranging (LiDAR) sensors. Each sensor type has unique characteristics and tradeoffs. In our work, we focus on widely available commodity ToF cameras.
% Different types of depth cameras are available based on different methods of acquiring depth data and processing it. Some of the most common types of depth sensors include stereo matching sensors, Time of Flight (ToF) sensors, and Light Detection and Ranging (LiDAR) sensors.
% The cameras built using each of these sensor types  have a set of unique characteristics which make them viable in different fields.
In particular, ToF cameras are low-cost devices that
% acquire the distance of the observed surface points by computing the round-trip time of a pulsating IR laser light\cite{hansard2012time}. These cameras 
provide a reasonable output depth resolution, a wide field of view, and a low level of noise \cite{kurillo2022azurev2}
% .Considering these characteristics along with the low cost, ToF cameras have been 
and are widely used in a range of different research topics from human-computer interaction (HCI)~\cite{liu2019kinecthci} to healthcare~\cite{pohlmann2016health}.

One of the main limitations of these types of depth cameras is the existence of latency and low output frame rate. For instance, Microsoft Kinect v2 comes with
% is able to output depth with a resolution of 512 $\times$ 424 within a range of 0.5–4.5 m, but only with
an acquisition rate of only 30 fps \cite{kurillo2022azurev2}. Additionally, Lu \etal \cite{lu2017hybrid} reported latency of up to 93 ms in Kinect v2. The Microsoft Azure Kinect Development Kit is a more recent version that incorporates the same technology for acquiring depth. Despite improving upon the previous version in several ways, \eg improving maximum resolution and depth range,
% \eg offering multiple spatial resolutions [max resolution of 1024 $\times$ 1024 for Wide field of view mode (WFOV) and 640 $\times$ 576 for narrow field of view mode (NFOV)], and wider depth ranges (0.25 to 2.88 m for WFOV and 0.50 to 3.86 m for NFOV),
it is still only able to achieve a maximum frame rate of 30 fps \cite{kurillo2022azurev2, kinect}. Our measurements suggest a latency of approximately 90 ms in the Azure Kinect output. The low output frame rate problem, combined with the high latency, imposes limitations on the applications of these cameras: fast motions are hard to capture accurately, and interactive applications suffer from input lag.

\begin{figure*}
\centering 
% \begin{subfigure}{0.68\linewidth} %
% \fbox{\rule{0pt}{2in}
\includegraphics[width=0.7\linewidth]{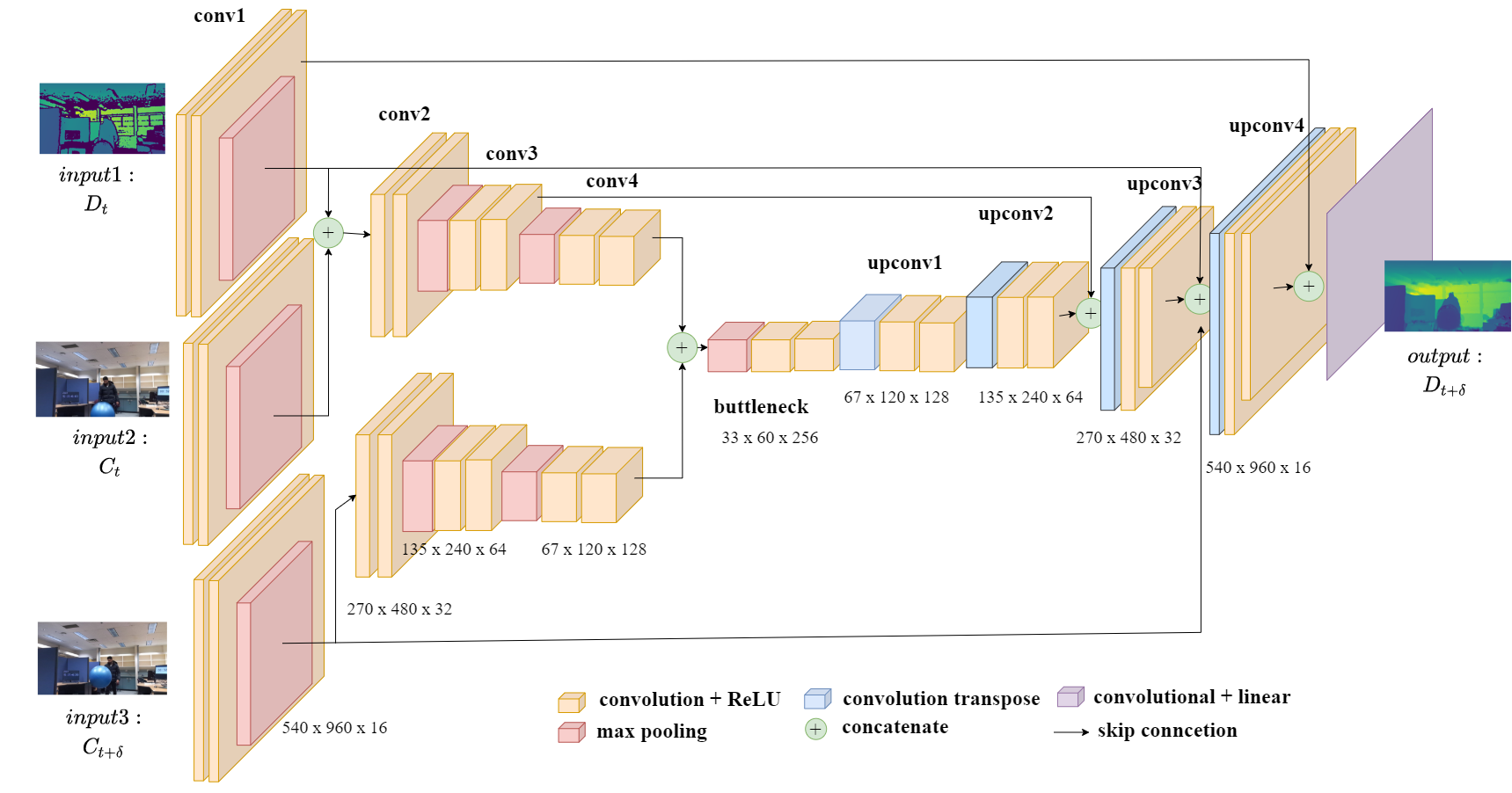}
% \caption{An example of a subfigure.}
% \label{fig:short-a} \end{subfigure} \hfill{}\begin{subfigure}{0.28\linewidth}
% \fbox{\rule{0pt}{2in} \rule{0.9\linewidth}{0pt}} \caption{Another example of a subfigure.}
% \label{fig:short-b} 
% \end{subfigure} 
\caption{AutoDepthNet is an encoder-decoder model that captures frames from a high-speed RGB camera ($C_t$, and $C_{t+\delta}$) and combines them with previous depth frames ($D_t$) to reconstruct future depth maps ($D_{t+\delta}$) with a high frame rate. }
\label{fig:network} 
\end{figure*}

On the other hand, in the past decade, advancements in the field of imaging have resulted in the creation of low-cost RGB cameras with both high-resolution and high speed. In fact, many consumer smartphones are now equipped with high-speed sensors in their cameras.
Several studies have tried to improve the output of depth cameras by taking advantage of RGB cameras. Most of these works have focused on depth super-resolution, \ie enhancing the spatial  resolution of depth maps~\cite{park2011upsling, riegler2016atgvupsample, hui2016sr}.
However, all these works assume that the RGB and depth video frame rates are identical.  Only a few other studies directly address the problem of temporal upsampling of depth cameras. These solutions are generally low resolution, hard to utilize with commodity depth cameras, or designed for a specific task, \eg object tracking. We will discuss details of these works in \cref{relatedworks}. 

Thus, the problem of low frame rate in depth cameras has still largely remained unexplored. In this study, we propose a framework for high accuracy and high frame rate reconstruction of depth maps, which ultimately can be used for temporal upsampling of depth cameras.  We take advantage of the fact that high-speed RGB cameras are ubiquitous and utilize them in our system for depth map reconstruction. Our framework utilizes a hybrid camera setup consisting of a ToF depth camera and a synchronized high frame rate RGB camera. We take advantage of the motion information in the RGB frame and update previous depth frames accordingly. We build and train an encoder-decoder model, called AutoDepthNet, for reconstructing depth.

The main contribution of this work is a system that reconstructs future depth maps with a high frame rate using a previous depth frame in combination with frames captured from a high-speed RGB camera. We are one of the first studies utilizing robust machine learning methods for temporal upsampling of depth maps.
% (that are typically used in MDE) %-------------------------------------------------------------------------

\section{Related Work}
\label{relatedworks}
% Several studies in the literature have addressed the subject of increasing frame rate and reducing the latency in depth cameras. The existing works differ in terms of the number and the types of cameras used for estimating the depth:

\subsection{Specialized Hardware Methods}
One approach is to build or modify the depth-sensing hardware to improve the frame rate. There are also several higher-speed depth sensors available in the market which use passive illumination and produce high-speed depth. However, passive illumination results in the creation of low-quality depth maps.  Kowdle \etal~\cite{kowdle2018need4} proposed a low-cost high-speed active-illumination depth sensor for depth estimation and obtained up to 800 fps in the low-resolution setting and 210 fps at high resolution. Even though their results are promising, incorporating these sensors with existing off-the-shelf depth cameras is challenging and is not possible for many consumers. Stuhmer \etal~\cite{stuhmer2015model} modified Kinect v2 to capture raw infrared images and performed model-based
tracking on the raw captures. While they obtained 300 fps for tracking objects, their method requires creating
models for both object and motion and is limited to only simple rigid shapes with simple motions.

\subsection{Hybrid Camera Methods}

% Using a hybrid camera setup is a common approach. While most consumer depth cameras, including Kinect v2 and Kinect Azure incorpo use an RGB camera along with the depth sensor.
% Most of the works in the literature 

Lu \etal~\cite{lu2017hybrid} proposed an optical flow-based method with a hybrid camera setup. They generated high frame rate depth frames by warping the latest Kinect depth image with the dense optical flow of high frame rate RGB images in a small region of interest (ROI). The resulting flow field is used
to extrapolate Kinect depth images. Considering the simplicity of their method and also small ROIs, they achieve 500 fps depth with 20ms latency. They also presented results for simple tracking applications. However, the generated depth maps suffer from high noise, which limits their usefulness for precise tracking.
Yuan \etal~\cite{yuan2018temporal} use a hybrid camera setup and a more advanced scene flow estimation technique. Their optimization technique estimates scene flow and guides the interpolation of intermediate depth
maps. While they achieve a better RMSE, their method fails with high-speed motion or severe occlusion.

\subsection{Monocular Depth Estimation}

The goal of monocular depth estimation (MDE) is to estimate the depth value of each pixel given a single RGB image.
Most MDE works use generative adversarial networks (GAN) \cite{kwak2020novelgan, aleotti2018generative} as they need to estimate the depth directly from another space (RGB). State-of-the-art MDE implementations use deep and complex neural networks trained on large datasets, which suffer from high complexity and long inference time. While MDE is not directly related to the temporal upsampling of depth maps, we inspire from the SOTA MDE methods in our design for reconstructing depth.  Unlike MDE, in this study, we take advantage of having the actual previous depth frame as input and future depth frames as ground truth (gt).

We exploit using a hybrid camera setup to address this problem. Since optical flow methods are not robust enough to deal with challenging examples, we use a learning-based technique, but with a hybrid camera setup. Our overall network design is an encoder-decoder architecture, similar to \cite{wofk2019fastdepth}, but we update the network design and inputs for optimum depth reconstruction.

%-------------------------------------------------------------------------

%------------------------------------------------------------------------

\section{Methodology}

\label{sec:Methodology}
%-------------------------------------------------------------------------

\subsection{Dataset}
Several public RGB-D datasets are available for optical flow estimation, monocular depth estimation, \etc.\cite{lopes2022survey}. However, almost all of them provide low frame rate RGB data (ranging from 5 to 30 fps), making them unsuitable for training our system.
Therefore, we collect our own dataset for training the model. The dataset consists of 5 different sequences with 240 fps RGB data and 30 fps depth data.

The background is consistent among sequences 1-5 and consists of different office objects (desks, cabinets, monitors, \etc) located at different depths. Sequence 6 is collected using a different background and lighting condition. Each sequence starts with a series of calibration frames in which the subject holds a checkerboard in their hand and moves in front of the camera, followed by the actual scenario in which the subject moves and performs different gestures or holds and moves objects \eg a ball.

Sequences 1 and 2 are about 270 sec long and sequences 3, 4, and 5, and 6 are shorter at 150 sec duration but contain faster subject movement. Therefore, the whole dataset consists of a total of 1140 seconds of video comprising 34,200 depth frames and 273,600 RGB frames.

%-------------------------------------------------------------------------

\subsubsection{Hardware Setup} \label{hardware} 
\begin{figure}[t]
\centering %
\vspace{-0.5cm}
\begin{subfigure}[t]{0.64\linewidth}
    \centering 
    \includegraphics[width=0.9\linewidth]{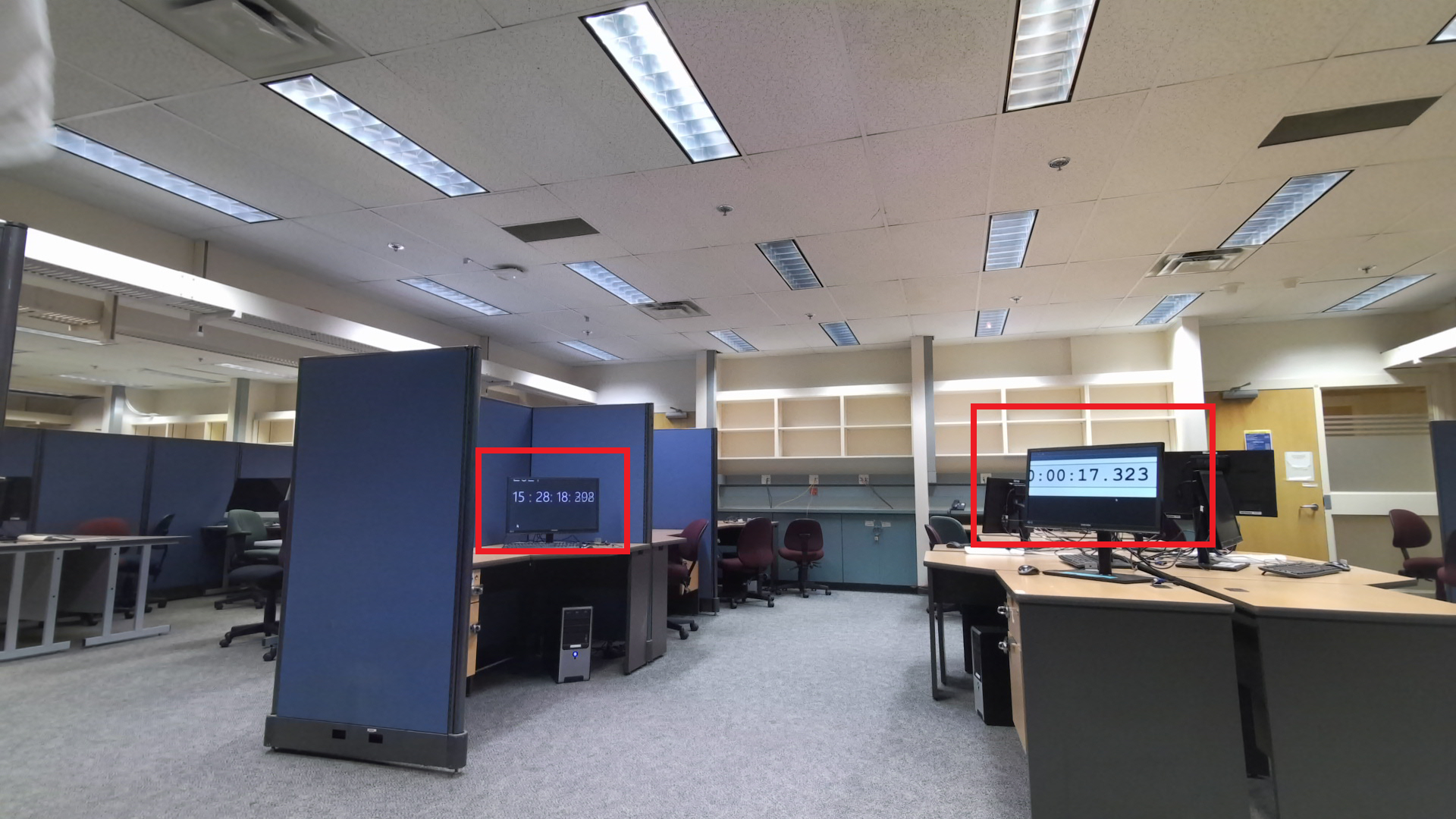} 
\caption{}
\end{subfigure}
\begin{subfigure}[t]{0.35\linewidth}
    \includegraphics[width=\linewidth]{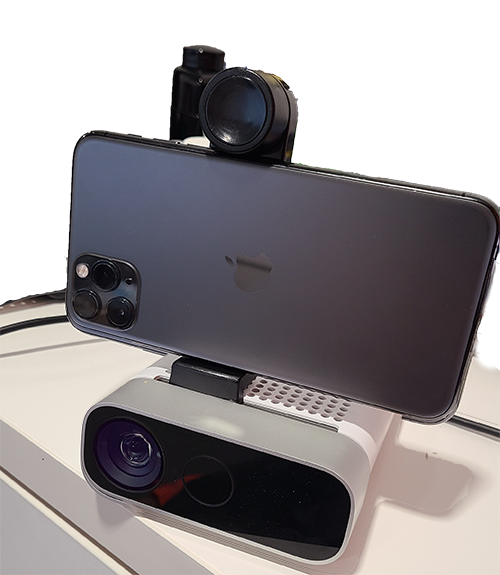} 
\caption{}
\label{fig:setup}
\end{subfigure}
\caption{Data collection Setup: (a) A sample frame from the RGB sequences. Two monitors (highlighted with red boxes) are placed in the background in order to provide visual timestamps for synchronizing the depth and RGB sequences. (b) Data collection camera setup.}
\end{figure}

We use a hybrid camera setup for our prototype. For a high frame rate RGB device, we use an iPhone 11 Pro camera recording in slow-motion video mode (240 fps, 1920 $\times$ 1080 pixels per frame). % that is used is an iPhone 11 Pro camera. While recording, the camera is set to the slow-motion video recording mode, which uses the wide lens camera (12 Megapixels) in 1080p resolution (image size of 1920 $\times$ 1080 for each RGB channel) and a frame rate of 240 fps.
For the depth camera, we use the Azure Kinect Development Kit, which has a 12-megapixel RGB camera and a one-megapixel infrared Time-of-Flight (ToF) depth camera. Depth sequences are captured in the narrow field-of-view (NFOV) unbinned depth mode (30 fps, 640 $\times$ 576 pixels per frame). The iPhone camera is mounted on top of the Azure Kinect as illustrated in  \cref{fig:setup}.

\subsubsection{Data Synchronization}
In order to synchronize both the RGB iPhone frames and Azure Kinect DK depth frames, we use visual timestamps while collecting the data, displayed on a pair of monitors in the data collection scene. The timestamps reported by each data collection device might not be highly accurate and could be affected by latency and clock drift from either device~\cite{lu2017hybrid}. However, in the case of this study, since we are using the system offline, we can synchronize the RGB and depth sequences by finding a pair of frames within the two sequences that display the same visual timestamp. The Kinect Azure sequences include depth, IR, and RGB images for every frame. We use RGB images from the Kinect and these timestamps to synchronize the beginning of the iPhone frame. While this method of synchronization works for our dataset, practical use of our system would need to incorporate synchronization logic between the two cameras.
% \begin{figure}[t]
% \centering %
% \includegraphics[width=0.6\linewidth]{50.png} 
% \caption{Data collection Setup: (a) A sample frame from the RGB sequences. Two monitors (highlighted with red boxes) are placed in the background in order to provide visual timestamps for synchronizing the depth and RGB sequences. (b)}
% \label{fig:monitors} 
% \end{figure}
\subsubsection{Data Calibration and Preprocessing}
As mentioned in \cref{hardware}, every sequence starts with a checkerboard calibration scenario. We use OpenCV camera calibration in order to estimate the intrinsic and extrinsic camera parameters \cite{zhang2000calib}. The depth maps in Kinect Azure are in the same image space as the IR images, so we use the IR frames for estimating depth camera parameters. Once we obtain the parameters for both cameras, we correct lens distortion according to the intrinsic parameters. We then convert the depth map into a 3D point cloud and project it into the iPhone image plane to align both cameras.

\begin{figure*}[t]
\centering %
\begin{subfigure}[t]{0.25\linewidth}
    \centering 
    \includegraphics[width=0.7\linewidth]{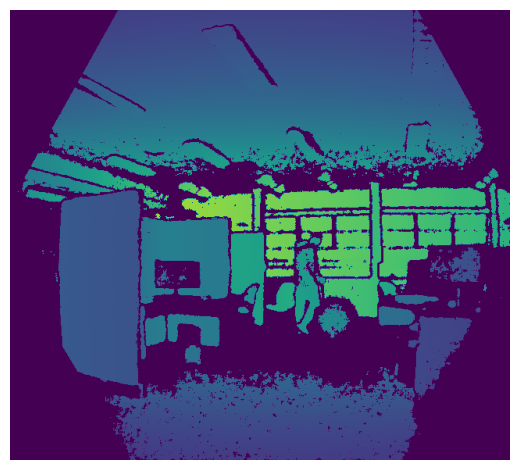}
   \caption{depth image before calibration}
   \label{fig:croporg} 
\end{subfigure}
\begin{subfigure}[t]{0.35\linewidth}
    \centering 
    \includegraphics[width=0.8\linewidth]{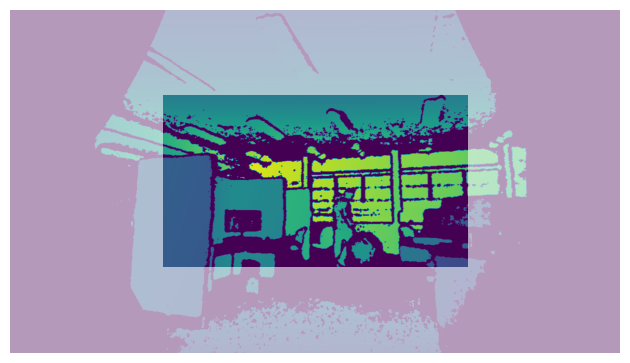}
   \caption{depth image after calibration}
   \label{fig:cropdep} 
\end{subfigure}
\begin{subfigure}[t]{0.35\linewidth}
    \centering 
    \includegraphics[width=0.8\linewidth]{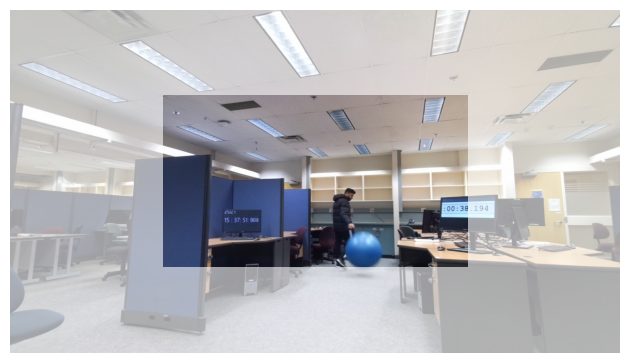}
  \caption{RGB image}
  \label{fig:croprgb} 
\end{subfigure}
\caption{\small A sample calibrated depth image with the cropped bounding box. The faded area is the original calibrated image and the non-faded part is the cropped section}
\label{fig:crop} 
\end{figure*}

After calibration, all RGB images will have a size of 1920 $\times$ 1080 $\times$ 3 and the depth images will have a size of 1920 $\times$ 1080 $\times$ 1. \cref{fig:cropdep} and \cref{fig:croprgb} show a sample of paired calibrated RGB and depth frames. Outside of a central area (highlighted), many pixels are invalid (outside the field-of-view of the depth camera) or stationary. Therefore, in order to speed up the network while keeping the important information in the image, we crop the calibrated images to half size from the center. We do not do any further processing or resizing on the data as the goal is to produce high-resolution depth images.
Therefore, the shape of the network's input is 960 $\times$ 540 $\times$ 1 for the depth and 960 $\times$ 540 $\times$ 3 for RGB images.

%-------------------------------------------------------------------------

\subsection{Network Design}
We use a convolutional encoder-decoder architecture for reconstructing depth maps. The network structure is shown in \cref{fig:network}. The overall network design is inspired by the U-net ~\cite{unet} architecture. We incorporate a separate encoder for the RGB input which is then merged with the depth encoder. The encoders extract high-level features from the previous frame's depth and RGB images ($D_t, C_t$), as well as the current frame's RGB image ($C_{t+\delta}$). The bottleneck applies several convolutions to the decoder's extracted latent vectors and passes them to the decoder part of the network.
The objective of the decoder part is to reconstruct the depth image by upsampling encoded features using transposed convolutional layers. 
% to avoid the checkerboard artifact~\cite{odena2016deconvolution}, we use a kernel size that is divisible by the stride.
All convolutional layers have ReLU activation. Finally, in order to match the output dimensions, at the last layer, we use a 1 $\times$ 1 convolutional layer with linear activation.
The network inputs are the following:

\begin{itemize}
\small
\itemsep0em 
\item input I: $C_{t}$, \qquad \quad RGB colour frame at time $t$
\item input II: $D_{t}$, \qquad \qquad depth frame at time $t$
\item input III: $C_{t+\delta}$, \quad RGB colour frame at time $t+\delta$
\end{itemize}
 During training, we use colour frames corresponding to the next available depth frame, resulting in a fixed $\delta$ of 1/30 seconds (33 ms). However, during inference, the colour frame can be chosen from any point in the future, effectively varying $\delta$.
The output of the network predicts:
\begin{itemize}
\item output: $D^{*}_{t+\delta}$, reconstructed depth map at time $t+\delta$
\end{itemize}
We use the next depth frame, \ie $D_{t+33}$ as the ground truth during training.
The network uses concatenative skip-connections for feature reusability and to leverage characteristics of early levels in deeper layers. We directly connect the features extracted from the inputs  $D_{t}$ and $C_{t+33}$ after two convolutional layers to the output. 

\subsection{Training Setup}
\label{loss}
%-------------------------------------------------------------------------
The Kinect camera incorporates logic to invalidate low-confidence or invalid depth pixels. These include pixels that are occluded or outside the IR illumination range, resulting in ambiguous depth measurements ~\cite{kinect}. Such invalid pixels are common, for example, on the edges between different objects, resulting in missing depth values at object borders.
On average, in our dataset, 29.56\% of all pixels are marked as invalid by the Kinect depth camera (indicated by a depth measurement of zero).
We suppress the effect of these pixels by ignoring them during training and evaluation. Therefore the final loss is obtained by calculating the Root Mean Squared Error (RMSE) between the reconstructed depth image and the gt over only the valid pixels in the gt image.
% \begin{equation}
% L_{zr}=\sqrt{\frac{\sum^N_{i,j=1}(x_{i,j}-X_{i,j})^{2}}{N}}; x\in D^{*}_{zr} \text{ and }  X\in D
% \end{equation}
% where $x_{i,j}$ is the pixel $(i,j)$ in $D^*_{zr}$, \ie the reconstructed depth image after removing the pixels from the gt image on it is defined as:

% \[
% D^{*}_{zr}(i,j) = \begin{cases} \phantom{} D^{*}(i,j) & \text{if } D_{t+30}(i,j) 
% \neq 0 \\ 0 & \text{if } D_{t+30}(i,j) = 0 \end{cases}
% \]
% and $X_{i,j}$ is the pixel $(i,j)$ in the gt frame ($D_{t+30}$).
% \subsection{Training Setup and Hyperparameters}
The network is trained end-to-end using the aforementioned loss on the training sequences 1-5. In order to avoid overfitting and to eliminate the temporal adjacency effect, we use leave-one-sequence-out cross-validation for evaluating the model performance, \ie training the model with 4 sequences at a time and using the entire length of the remaining sequence for testing. Several experiments were performed to fine-tune the hyperparameters.
% We trained the model for 50 epochs, batch size of 32, a learning rate of 0.001 with a 0.5 reduction rate on the plateau, and using the Adam optimizer \cite{kingma2014adam}. 

%------------------------------------------------------------------------
%-------------------------------------------------------------------------

\section{Experiments}

We compare the results with a naive baseline, consisting simply of using the previous depth frame as the predicted frame, as well as a comparison method based on optical flow and one of the state-of-the-art MDE methods, \ie BinsFormer with Swin-Large encoder\cite{li2022binsformer}. We use a simplified version of the Lu \etal ~\cite{lu2017hybrid} technique. First, the dense optical flow between the RGB images of the target frame and the original frame is computed using Farneback's algorithm \cite{farneback2003flow}. Then, we warp the input depth map using the obtained dense optical flow.

\newcommand{\specialcell}[2][c]{%
  \begin{tabular}[#1]{@{}c@{}}#2\end{tabular}}
 
%  \begin{table*}
%   \centering
%  \begin{tabular}[c]{l|c|c|c}
%  \hline 
%  \toprule
% {\# of sequence} &\specialcell{ RMSE without\\ removing 0 pixels} & \specialcell{RMSE 0 removed \\ from gt and input} & \specialcell{RMSE gt 0 values removed \\from the input only}\tabularnewline

% \midrule
% 1 & 1559.3309  $\pm$ 317.8329 &  125.41126  $\pm$ 116.9843 & 1108.0605  $\pm$ 226.37753\tabularnewline

% 2 & 1640.8559 $\pm$ 296.1815 & 140.1249 $\pm$ 138.9423  & 1209.3830 $\pm$ 185.4116 \tabularnewline

% 3 & 1759.3772 $\pm$ 247.3009  &  129.4181 $\pm$ 135.0207 & 1251.2139 $\pm$ 178.5969\tabularnewline

% 4 & 1655.5327 $\pm$ 185.185 & 182.25894 $\pm$ 146.4124 &  1181.7269 $\pm$ 140.9978\tabularnewline

% 5 & 1711.3466 $\pm$ 289.1173 & 127.0832 $\pm$ 139.1450 & 1206.6603  $\pm$ 201.7221 \tabularnewline

% no movement & 316.1822 $\pm$ 15.1705 & 4.42580 $\pm$ 1.5995 & 222.9787 $\pm$ 19.0664  \tabularnewline
% \bottomrule
% \end{tabular}
%   \caption{Baseline RMSE values for different sequences in the dataset,  without removing invalid 0 pixels, with removing invalid pixels from both, and with removing the 0 pixels in the gt from the input image. The reported values are mean $\pm$ std in mm}
%   \label{tab:baseline}
% \end{table*}
\subsection{Depth Reconstruction Results}
\begin{figure*}
\centering 
% \begin{subfigure}{0.68\linewidth} %
% \fbox{\rule{0pt}{2in}
\vspace{-0.5cm}
\begin{subfigure}{0.33\linewidth}
    \centering 
    \includegraphics[width=0.8\linewidth, trim={0 1cm 0 0 },clip]{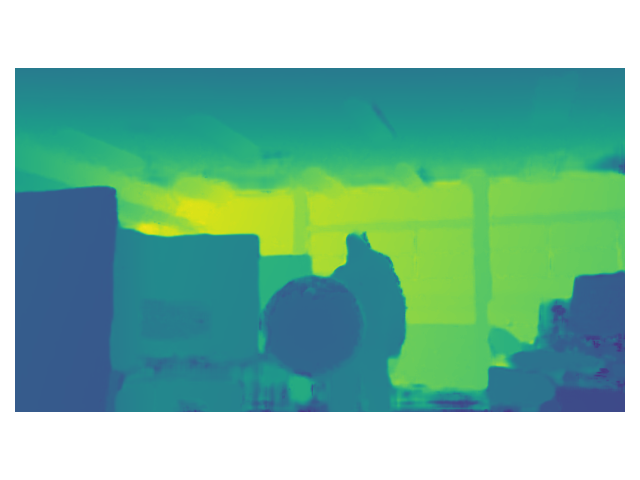}
    \vspace{-0.1cm}
    % \vspace{-0.5cm}
    \caption{reconstructed image}
\end{subfigure}
\begin{subfigure}{0.33\linewidth}
    \centering 
    \includegraphics[width=0.8\linewidth, trim={0 1cm 0 0 },clip]{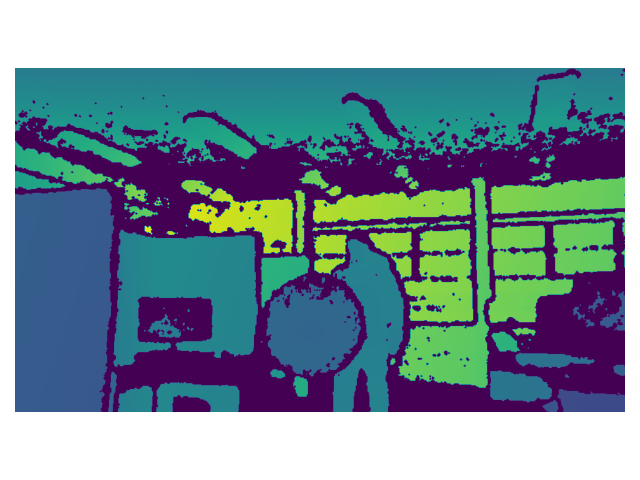}
    \vspace{-0.1cm}
    \caption{gt image}
\end{subfigure}
% \begin{subfigure}{0.49\linewidth}
%     \centering 
% \includegraphics[width=0.6\linewidth]{\string"pred_zeropained.png"}
% \vspace{-0.5cm} \caption{reconstructed image with adding invalided pixels mask from gt}
%     \end{subfigure}
\begin{subfigure}{0.33\linewidth}
    \centering 
\includegraphics[width=0.8\linewidth, trim={0 1cm 0 0 },clip]{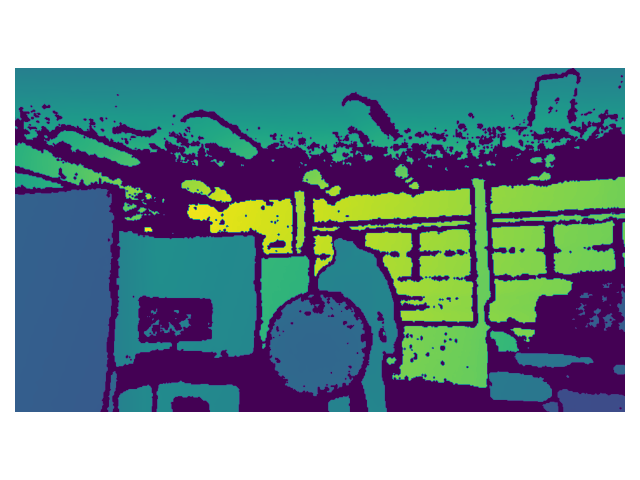}
\vspace{-0.1cm}
% \vspace{-0.4cm} 
\caption{input depth image}
    \end{subfigure}
% \caption{An example of a subfigure.}
% \label{fig:short-a} \end{subfigure} \hfill{}\begin{subfigure}{0.28\linewidth}
% \fbox{\rule{0pt}{2in} \rule{0.9\linewidth}{0pt}} \caption{Another example of a subfigure.}
% \label{fig:short-b} 
% \end{subfigure} 

\caption{\small Results of the network reconstructed depth maps}
\label{fig:pred} 
\end{figure*}

We present the results of the depth map reconstruction and compare them with the baseline and the optical flow method. \cref{fig:pred} shows the reconstructed depth image for a sample from sequence 3 as well as the gt image and the input frame. \cref{tab:rmse} presents the RMSE values between the reconstructed image and gt, baseline, optical flow method, and MDE. 
% In order to calculate the RMSE values, we remove the invalid pixels from both the input frame and the gt.
The network is able to take advantage of the motion in the RGB image and use it to reconstruct the depth frame. The results show that the reconstructed image outperforms the optical flow method by 45.1\%, reducing the average RMSE by 0.061. The results also show that AutoDepthNet achieves a much smaller RMSE compared to Binsformer \cite{li2022binsformer}, which is one of the SOTA MDE methods. This was expected because MDE addresses a harder problem and unlike our method, it does not take the advantage of using previous depth frames.

 \begin{table*}
 \small
  \centering
    \begin{tabular}[c]{l|c|c|c|c|c|c|c}
    \hline 
    \toprule
    \backslashbox{method}{seq.}&1&2&3&4&5&6&avg.
 \tabularnewline

\midrule
Baseline & 0.125  &0.140   & 0.129 & 0.182&0.127&0.152  &0.142\tabularnewline
% \midrule
Optical flow & 0.120 &  0.137 & 0.128 &0.170&0.125&0.142 &0.137\tabularnewline
% \midrule
Binsformer & 0.269 & 0.284  & 0.275 & 0.313 &0.291&0.284& 0.286\tabularnewline
% \midrule
Ours &\textbf{0.069}& \textbf{0.076}&\textbf{0.072}&\textbf{0.085} & \textbf{0.070} & \textbf{0.087} & \textbf{0.076}\tabularnewline
% \midrule
\bottomrule
\end{tabular}
  \caption{Depth reconstruction results. The reported values are the RMSE between each method's output and the gt among different test sequences in the dataset. The baseline corresponds to the RMSE between the previous depth frame and gt (after removing the invalid pixels from both). For MDE we used Binsformer\cite{li2022binsformer} method.}
   \label{tab:rmse}
\end{table*}

%  \begin{table}
%  \small
%   \centering
%     \begin{tabular}[c]{l|c|c|c|c}
%     \hline 
%     \toprule
%     \backslashbox{method}{seq.}
%  &\specialcell{Ours} & \specialcell{Baseline} & \specialcell{Optical flow} & \specialcell{MDE} \tabularnewline

% \midrule
% 1 & \textbf{0.069 } &0.125  & 0.120 & 0.269  \tabularnewline
% % \midrule
% 2 & \textbf{0.076} &  0.14  &0.137 & 0.284 \tabularnewline
% % \midrule
% 3 & \textbf{0.072} & 0.129  & 0.128 & 0.275\tabularnewline
% % \midrule
% 4 & \textbf{0.085} & 0.182  & 0.017 & 0.313\tabularnewline
% % \midrule
% 5 & \textbf{0.070 } & 0.127 & 0.125 & 0.291\tabularnewline
% % \midrule
% avg. & \textbf{0.074} & 0.140 & 0.136 & 0.286
% \bottomrule
% \end{tabular}
%   \caption{Depth reconstruction results. The reported values are the RMSE between each method's output and the gt among different test sequences in the dataset. The baseline corresponds to the RMSE between the previous depth frame and gt (after removing the invalid pixels from both). For MDE we used \cite{li2022binsformer} method}
%    \label{tab:rmse}
% \end{table}
\subsection{Model Time Consumption}
In addition to reconstructing accurate depth maps, one of the main goals of this work is to reduce the latency in depth cameras. Our testing showed an average latency of 89.7 ms in the Kinect camera feed. 
% many applications and future directions 
% Therefore, it is critical that the proposed framework be able to run in real time.
Inference time was done using TensorFlow profiling. We experimented with the model on 3 different hardware setups, including  one CPU device (Intel Core i7-10510U CPU) and two GPU devices: Nvidia Tesla V100 ``\textit{GPU I}'' and GeForce RTX 2060 ``\textit{GPU II}".

The first strategy to reduce inference time is to use smaller input sizes. We resized the input images down by a factor of 2, resulting in input images with a size of 480 $\times$ 270. We then re-scaled the predicted depth maps back to the original size of  960 $\times$ 540 using nearest-neighbor interpolation and then applying the invalid pixel masks. The results in \cref{tab:time} indicate a considerable 59.4\% reduction in time consumption while maintaining the overall performance (RMSE increased by only 0.007: 0.076 to 0.083). 
% This strategy can also be utilized by predicting the depth over certain regions of interest (ROI) to even further reduce the time consumption. For instance, in applications such as object tracking, we can use ROIs to predict the depth around the object that is being tracked a very high-speed depth reconstruction.

In addition, the profiling results indicated that the convolutional operations consume the majority of the inference time (38\%). To reduce the time complexity of the convolution operations, we replaced normal convolutions in the model with separable convolutions (S-Conv2D).
% which include a  depth-wise convolution followed by a point-wise convolution. The depth-wise convolution performs spatial convolution independently over each channel of an input and the point-wise convolution projects output channels onto a new channel space. 
The results in \cref{tab:time} show that on average, using separable convolutions results in an 18.5\% reduction in the model time consumption while the model performance was only reduced by 3\% (average RMSE increased from 0.076 to 0.078)

The results in \cref{tab:time} suggest that on average consumer hardware, \ie GPU II, the model can predict depth in 22.3 msec. On more powerful hardware (GPU I), the model has the capability to achieve more than 60 fps on the original input size and up to 200 fps on resized input with a latency of 8 ms using parallel processing (running pipeline steps on one frame in parallel with the model on a previous frame). The GPU II inference time leads to a depth reconstruction of over 78 fps, which is still substantially higher than the original Kinect camera frame rate (30 fps). Such a high frame rate, in addition to reducing latency, proves the framework's ability to be used for high-speed depth reconstruction.

 \begin{table}
 \small
%  \label{tab:time}
  \centering
 \begin{tabular}[c]{l|c|c|c|c|c|c} 
 \hline 
 \toprule
\specialcell{} &\multicolumn{2}{c}{\specialcell{Conv2D \\960 $\times$ 540}}  &\multicolumn{2}{c}{\specialcell{ S-Conv2D\\960 $\times$ 540}}  &\multicolumn{2}{c}{\specialcell{ S-Conv2D\\
480 $\times$ 270}}
 \tabularnewline
Device  &  {model} &  {total} &  {model} &  {total} &  {model} &  {total} \tabularnewline
\midrule
CPU  &  91.9 &  97.2   & 81.2 & 85.1   & 47.4  & 50.2
\tabularnewline
GPU I &   14.4 & 24.1 & 12.7 & 24.2 & 4.9 & 8.0
\tabularnewline
GPU II &  27.7 & 34.0 & 22.3 & 29.2 & 10.8 & 12.8
 \tabularnewline
\bottomrule
\end{tabular}
  \caption{\small Model time consumption results (in msec) on different hardware: CPU, GPU I, and GPU II. We report the inference time  for the model only and also the full inference pipeline, which includes the whole time spent between two consecutive frames (including the time to take the input, pass through the model and output the reconstructed depth)}
  \label{tab:time}
\end{table}

% \subsection{Model components}
\vspace{-0.2cm}
\subsection{Ablation 1: Encoder-Decoder Steps}
Using fewer cascades in the auto-encoder will result in fewer parameters and a smaller network. We test the effect of using a different number of cascades in the encoder and decoder. While keeping the overall network structure, we adjust the number of upsampling and downsampling steps to 2, 3, 4, and 5. With every step, we also increased the number of filters used within the convolutional layers.
% \cref{tab:cascades} presents the RMSE results for the different number of cascades and the number of parameters.
The experimental results show that by increasing the number of cascades from 2 to 4, the average RMSE prediction improved from 0.189 to 0.076, while the number of parameters also increased (186,625 to 2,904,001 parameters). We observed that however, using 5 cascades does not improve the performance but rather makes it slightly worse (RMSE of 0.081). A possible reason for this is that using too many pooling layers in the encoder part results in a small latent feature vector in the bottleneck (16 $\times$ 30), and the upsampling layers in the decoder would not be able to reconstruct the original frame from the small encoded image effectively.

%  \begin{table*}[h!]
%   \centering
%  \begin{tabular}[ht]{l|c|c|c}
%  \hline 
%  \toprule
% {\# of cascades} &\specialcell{ RMSE prediction and gt} &\specialcell{ computational cost (sec)} &\specialcell{ \# of parameters}\tabularnewline

% \midrule
% 2  & 189.1638 $\pm$ 72.4785 & 0.3249 & 186,625 \tabularnewline
% 3 &  131.2671  $\pm$ 80.1387 & 0.6280 & 723,777\tabularnewline

% 4 &   \textbf{74.8253  $\pm$ 33.9588}  & 0.3472 & 2,904,001 \tabularnewline
% 5 & 81.3673 $\pm$ 46.4490 & 0.9427& 11,491,521\tabularnewline
% \bottomrule
% \end{tabular}
%   \caption{Results of the effect of the number of cascades in the encoder-decoder. The reported values are the RMSE results between the reconstructed image and gt averaged over 5 sequences (in mm), the reported values for computational cost are by running the model on CPU.  }
%   \label{tab:cascades}
% \end{table*}

%  \begin{table}[h!]
%   \centering
%  \begin{tabular}[ht]{l|c|c}
%  \hline 
%  \toprule
% {\# of cascades} &\specialcell{ RMSE pred \& gt} &\specialcell{ \# of parameters}\tabularnewline

% \midrule
% 2  & 189.16 $\pm$ 72.47  & 186,625 \tabularnewline
% 3 &  131.26  $\pm$ 80.13  & 723,777\tabularnewline

% 4 &   \textbf{74.82  $\pm$ 33.95}  & 2,904,001 \tabularnewline
% 5 & 81.36 $\pm$ 46.44 & 11,491,521\tabularnewline
% \bottomrule
% \end{tabular}
%   \caption{Effect of number of cascades in the encoder-decoder. The reported values are the RMSE results between the reconstructed image and gt averaged over 5 sequences (in mm). }
%   \label{tab:cascades}
% \end{table}

\subsection{Ablation 2: Effect of Skip Connections}
We use skip-connections  to leverage  different network levels characteristics in deeper layers and to make use of scene changes in future frames while preserving the ability to reconstruct depth maps  that are close to input frames. We experimented with skip-connections between the initial and final layers.
% The hypothesis was that by connecting the final layers with RGB input, the network would learn to make essential changes and use scene changes that are presented in future frames.
% On the other hand, by connecting the final layers with the input depth image, the network sustains the ability to reconstruct meaningful depth maps that are closely tied to the Kinect output. 
As \cref{fig:network} shows, there are 4 skip-connections  in the network.
% : skip-connection between depth conv1 layer and upconv4 (D input connection), skip-connection between RGB pooling layer and upconv3 (RGB connection), skip-connection between the depth (D pooling connection), skip-connection between conv3 and upconv2 (D intermediate connection). 
We removed one skip-connection at a time for each experiment.  The results suggest that by keeping all the skip-connections, the model is able to achieve the best performance with an RMSE of 0.076. Moreover, removing the input depth skip-connection resulted in a higher RMSE value (0.169) compared to other connections suggesting that the network needs to use lower levels of depth information for reconstructing valid depth frames.
%  \begin{table}[h!]
%   \centering
%  \begin{tabular}[c]{l|c}
%  \hline 
%  \toprule
% {skip conncetion removed} &\specialcell{ RMSE pred \& gt} \tabularnewline

% \midrule
% No connection removed  &  \textbf{85.47  $\pm$ 34.07} \tabularnewline
% D input connection &  167.31  $\pm$ 30.12\tabularnewline

% RGB connection &   128.14 $\pm$ 47.30 \tabularnewline

% D intermediate connection & 109.20  $\pm$ 42.28 \tabularnewline

% RGB \& D input connections &  179.77 $\pm$ 27.39 \tabularnewline
% Baseline &  182.25 $\pm$ 146.41 \tabularnewline
% \bottomrule
% \end{tabular}
%   \caption{Results of the network's performance while removing skip-connections. The reported values are the RMSE results between the reconstructed image and the gt over sequences 4 (in mm) }
%   \label{tab:skip}
% \end{table}

\subsection{Increasing \bm{$\delta$}}
We experiment with the extent to which the network is able to reconstruct a depth map as $\delta$ increases. We provide RGB frames further in the future during inference and report the RMSE results, compared to the corresponding future depth frames. As \cref{tab:past} shows, the network reconstruction is better than the RMSE between the input and gt in all instances. The results also suggest that when using $t+67$ ms as input, \ie a two-frame gap between the depth and new RGB frame, the model still reconstructs a better depth than the original baseline (input t-1 and gt). These results also suggest that our approach could correct for up to 67 ms of Kinect latency without significantly compromising accuracy. As the network introduces as little as 8 ms of latency, the net latency reduction could be up to 59 ms.

\label{past}
 \begin{table}
 \small
  \centering
 \begin{tabular}[c]{l|c|c|c} 
 \hline 
 \toprule
\specialcell{$\delta$ \\value} &\specialcell{\# past\\ frames}&\specialcell{ RMSE \\prediction \& gt}  &\specialcell{ RMSE \\input \& gt}\tabularnewline

\midrule
33 ms & 1  &  \textbf{0.076  $\pm$ 0.033} &  0.142 $\pm$ 0.134 \tabularnewline
67 ms & 2  &  0.132  $\pm$ 0.094 & 0.224 $\pm$ 0.208\tabularnewline
100 ms & 3  &  0.212  $\pm$ 0.160 & 0.473 $\pm$ 0.236 \tabularnewline
\bottomrule
\end{tabular}
  \caption{Results of experimenting with the network's ability to reconstruct depth as $\delta$ increases, averaged over 6 seq.}
  \label{tab:past}
\end{table}

%------------------------------------------------------------------------

\section{Applications}
The proposed framework has the potential to extend and improve the use of existing depth cameras in many ways. We demonstrate applications of our method in body tracking, and video object segmentation (VOS), and also discuss other potential areas in which such a high frame rate technique can be utilized.

\subsection{Body Tracking}
\label{track}
The goal of tracking is to find the pose parameters that describe the pose of the object of interest at different times according to the data. A common tracking optimization is to update the pose from a prior frame. However, when the object moves rapidly, the pose of the subsequent frames can differ widely from that of the previous frame resulting in a loss of tracking. To address this, tracking systems typically must expensively reinitialize the pose or implement much more complex tracking schemes to compensate \cite{kowdle2018need4}. However, high frame rate data will effectively reduce inter-frame motions and result in more efficient tracking systems.

Generally, human body tracking and pose estimation 
% \ie  the task of predicting locations of body joints from input frames, 
is performed either in 2D or 3D. 3D pose estimation is more challenging as it needs to predict the depth of body joints in addition to the x and y coordinates. Direct 3D pose estimation systems in the literature typically employ complex and slow deep neural network architectures \cite{chen2020monocular} that are not suitable for real-time use.  
Therefore, for faster tracking, indirect 3D pose estimation techniques first predict joint locations from the RGB image in the 2D surface and then extend them to 3D space.
% They can be classified as Regression-based, detection-based, top-down, and bottom-up approaches \cite{chen2020monocular}. 
We integrate two widely used pose estimation solutions with our framework and demonstrate the robustness of the proposed method in such applications. 

\begin{figure}
\centering 

% \begin{subfigure}{0.68\linewidth} %
% \fbox{\rule{0pt}{2in}
\begin{subfigure}[t]{0.8\linewidth}
    \centering 
    \includegraphics[width=1\linewidth]{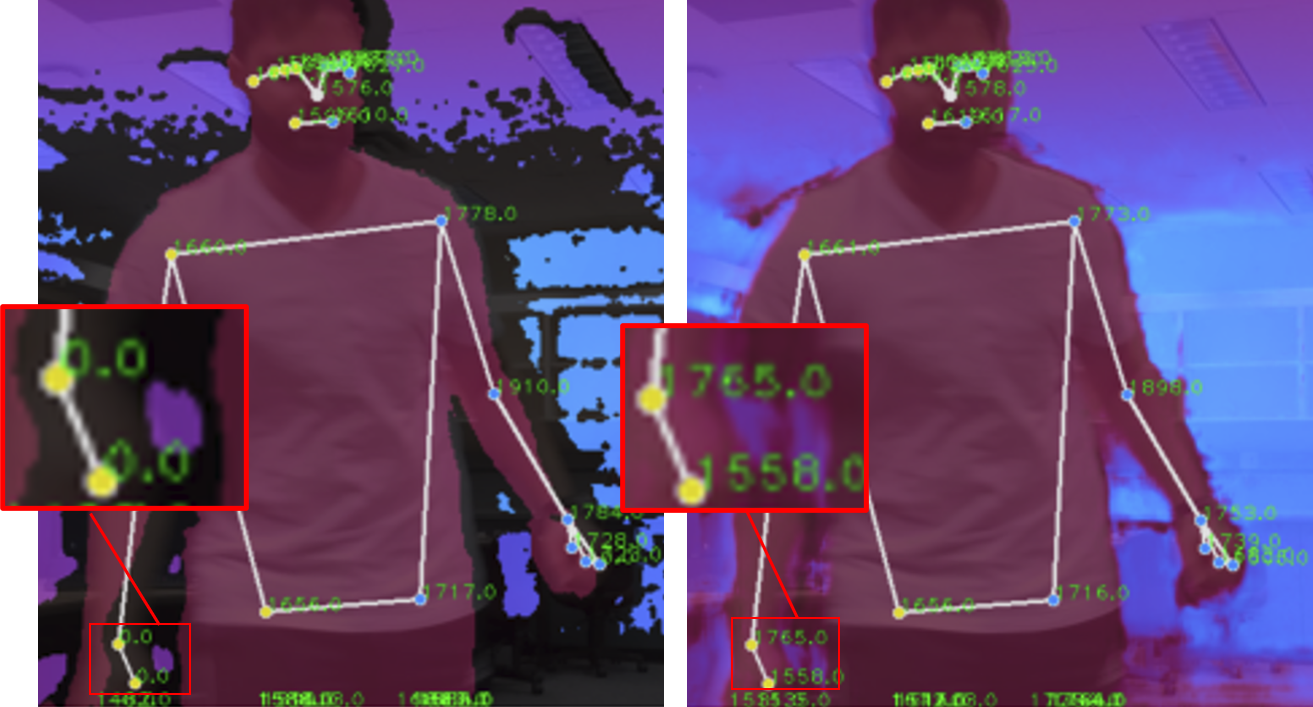}
    \caption{MediaPipe body tracking integrated with Kinect depth map (left) and AutoDepthNet (right) }
\end{subfigure}
\begin{subfigure}[t]{0.8\linewidth}
    \centering 
    \includegraphics[width=0.98\linewidth]{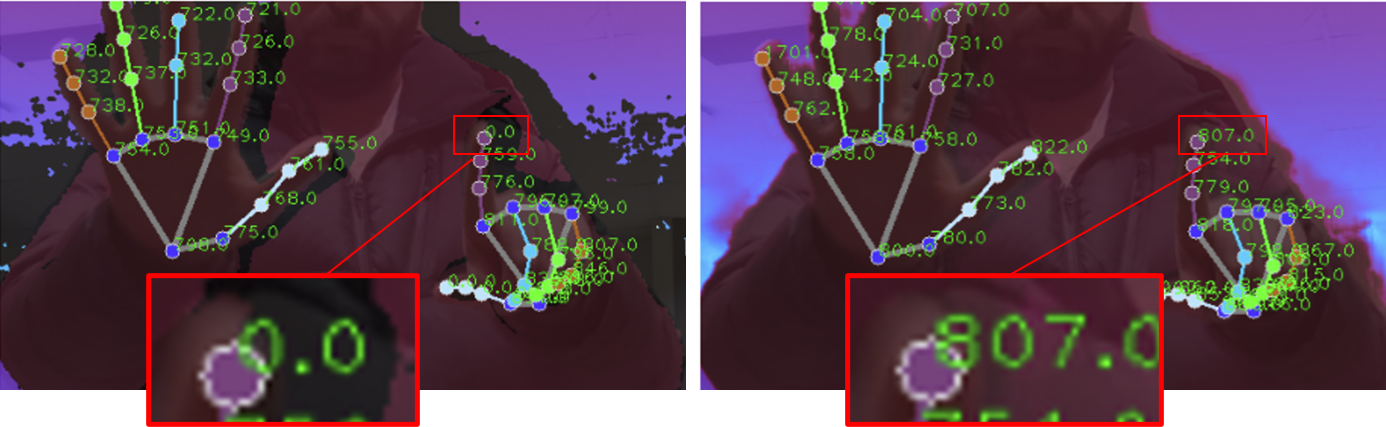}
    \caption{MediaPipe hand tracking integrated with Kinect depth map (left) and AutoDepthNet (right)}
\end{subfigure}
% \begin{subfigure}[t]{0.47\linewidth}
%     \centering 
%     \includegraphics[width=1\linewidth]{\string"medigthand.png"}
%     \caption{MediaPipe hand tracking integrated with Kinect  depth map}
% \end{subfigure}
% \begin{subfigure}[t]{0.47\linewidth}
%     \centering 
%     \includegraphics[width=1\linewidth]{\string"medihand.png"}
%     \caption{MediaPipe hand tracking integrated with AutoDepthNet}
% \end{subfigure}
% \caption{An example of a subfigure.}
% \label{fig:short-a} \end{subfigure} \hfill{}\begin{subfigure}{0.28\linewidth}
% \fbox{\rule{0pt}{2in} \rule{0.9\linewidth}{0pt}} \caption{Another example of a subfigure.}
% \label{fig:short-b} 
% \end{subfigure} 
\caption{\small Integrating MediaPipe body and hand tracking solutions with AutoDepthNet. Depth values are shown in green on the images. Zoomed-in areas show examples of situations where gt depth fails to provide depth value for some of the landmarks, while AutoDepthNet's reconstructed depth values are consistent.}
\label{fig:media} 
\end{figure}

Google's MediaPipe solution \cite{lugaresi2019mediapipe,zhang2020mediapipe} consists of multiple models for detecting bounding boxes and predicting the hand and body skeletons and is able to obtain 2.5 D landmarks in real time. However, the depth values are not accurate and are estimated using the relative depth with reference to the hips (for body tracking) or wrist (for hand tracking) position. Therefore, we update the depth values using our reconstruction and build a fast 3D pose estimation. We use MediaPipe to extract 2D landmarks from high-speed RGB frames and then 
% The current RGB frame and also previous depth frames are passed through AutoDepthNet. 
update the depth value of every pixel using the reconstructed depth map from AutoDepthNet. In addition to reducing the latency and providing the depth values at a faster frame rate and ultimately capturing fast movements, AutoDepthNet provides consistent depth values for every single joint, while the Kinect depth camera fails in some cases. \cref{fig:media} demonstrates MediaPipe and AutoDepthNet's fused 3D body tracking. 

\begin{figure}
\centering 
% \begin{subfigure}{0.68\linewidth} %
% \fbox{\rule{0pt}{2in}
\begin{subfigure}{0.49\linewidth}
    \centering 
    \includegraphics[width=0.75\linewidth]{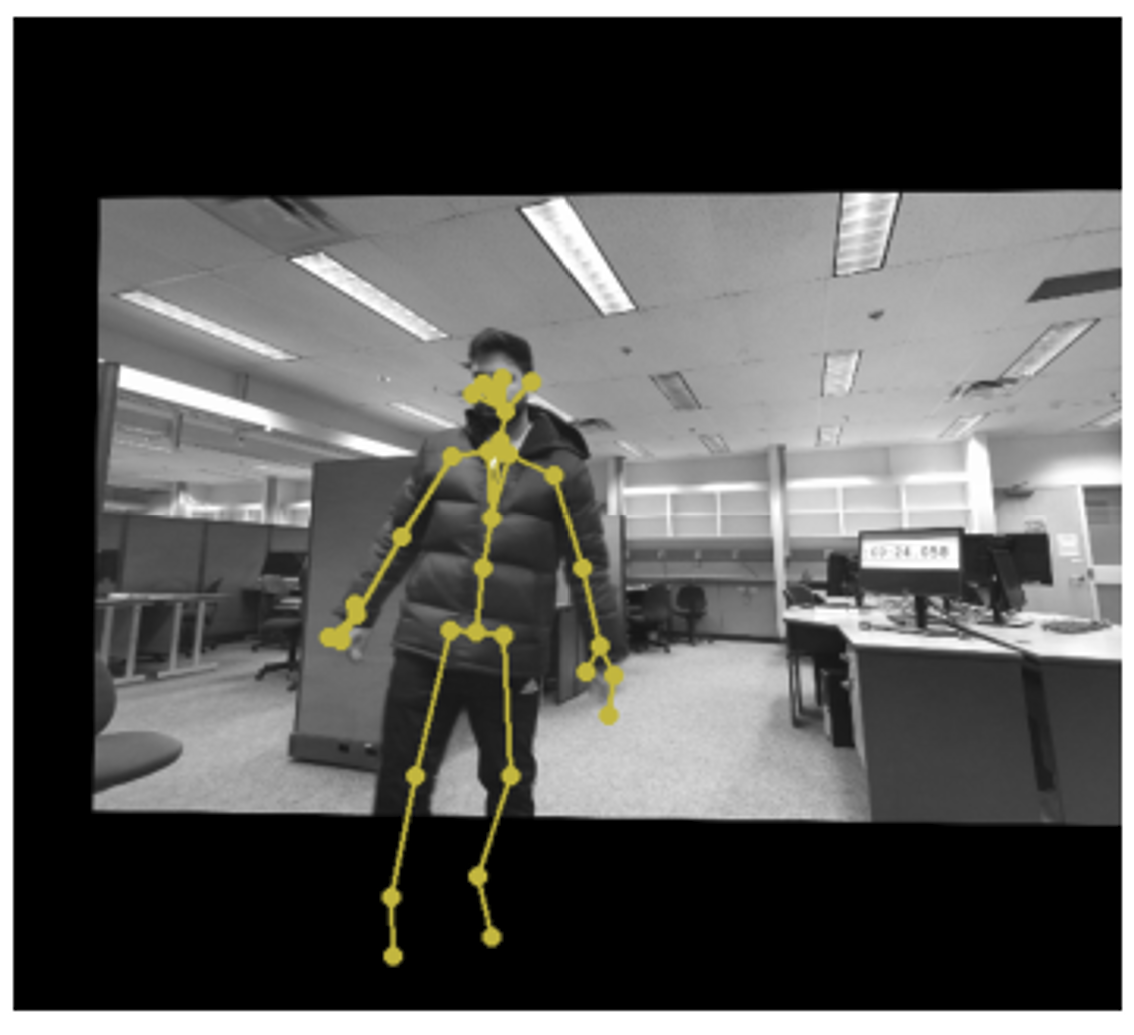}
    \caption{Synthesized IR image}
\end{subfigure}
\begin{subfigure}{0.49\linewidth}
    \centering 
    \includegraphics[width=.75\linewidth]{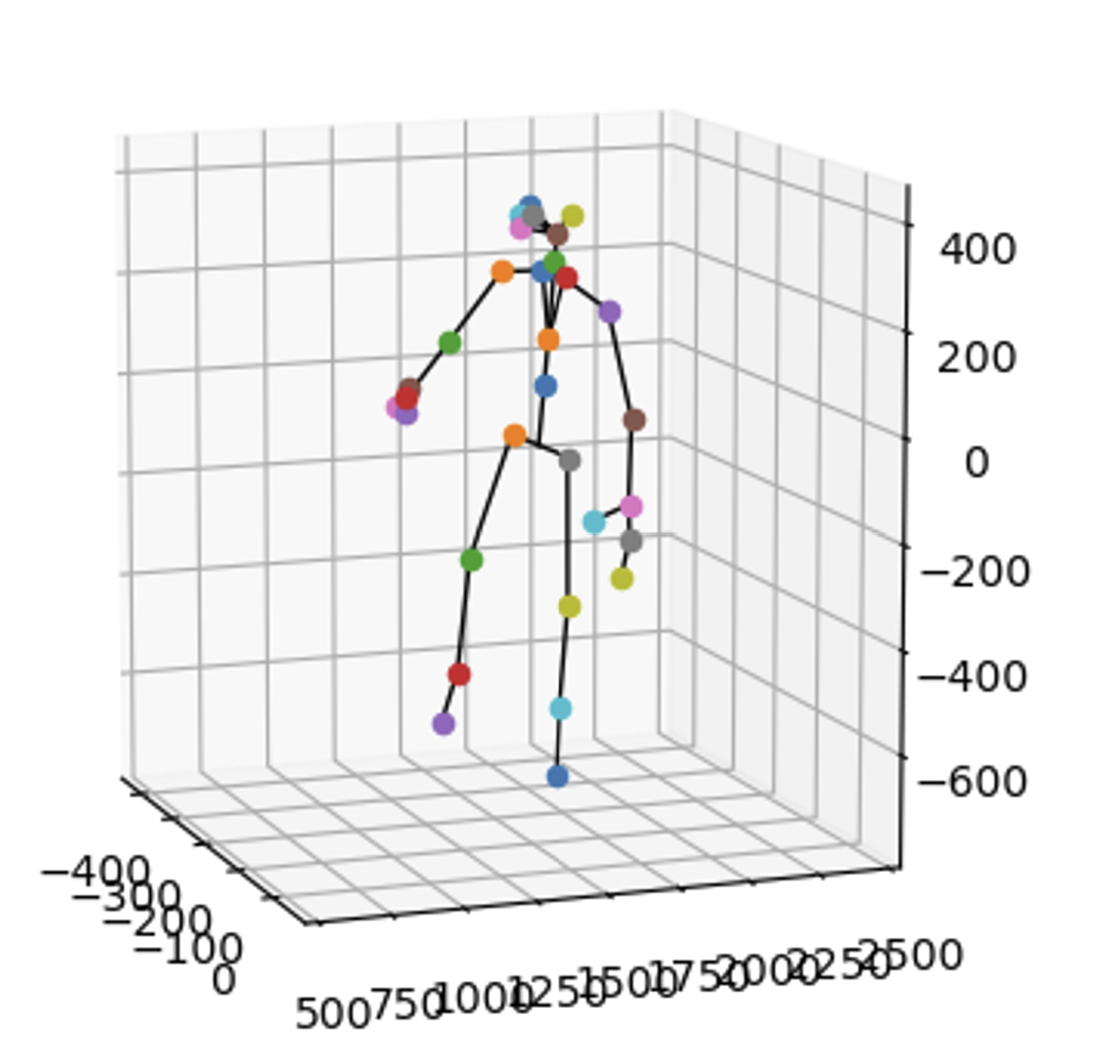}
    \caption{3D visualization of joints}
\end{subfigure}
% \begin{subfigure}{0.33\linewidth}
%     \centering 
%     \includegraphics[width=0.6\linewidth]{\string"bl2.png"}
%     \caption{3D model in Blender app}
%     \label{fig:mediac}
% \end{subfigure}

% \caption{An example of a subfigure.}
% \label{fig:short-a} \end{subfigure} \hfill{}\begin{subfigure}{0.28\linewidth}
% \fbox{\rule{0pt}{2in} \rule{0.9\linewidth}{0pt}} \caption{Another example of a subfigure.}
% \label{fig:short-b} 
% \end{subfigure} 
\caption{Integrating AKBT with AutoDepthNet}
\label{fig:akbt} 
\end{figure}

Unlike MediaPipe, Azure Kinect Body Tracking (AKBT) \cite{kinect}, provided with the Azure Kinect SDK, directly utilizes RGB, IR, and depth frames and offers 3D pose estimation. Despite the good performance \cite{romeo2021performance}, this solution is also confined by the high latency and low frame rate of the depth camera.
% We integrate our framework with the AKBT pipeline to obtain high frame rate depth maps. 
The model contains a CNN which obtains 2D key points from the IR image and then the 3D model fitting is done using the depth maps to output 3D joint locations. To run the model with a high frame rate, we synthesize ``IR'' images from RGB by extracting the luminance for every high fps RGB frame and then warping it into the original IR image dimensions and feeding it to the model. We also replace the depth images with AutoDepthNet's reconstructed depth and finally obtain 3D poses. \cref{fig:akbt} shows pose estimation using a synthetic IR image and  the corresponding 3D visualization of joints. 
% Using the 3D poses and integrating them with a 3D model which can be ultimately used in order to perform high speed body and hand tracking on VR models. We use Blender \cite{blender} to create the 3D model and transfer tracking animations to the model as demonstrated in \cref{fig:mediac}.

\subsection{Video Object Segmentation}
The proposed framework's overall design can be used for upsampling different types of hybrid sparse sequences/datasets. We demonstrate this by showcasing the effectiveness of using AutoDepthNet for reconstructing Video Object Segmentation (VOS) sequences. We updated the model's final layers and replaced RMSE loss with Jaccard Index (J). We retrained AutoDepthNet on the YouTube-VOS dataset \cite{xu2018youtube}.  $C_{t}$ inputs are simply replaced with every RGB frame in the dataset and $D_{t}$ inputs correspond to every 5th annotation frame. During inference, our model uses previous annotations and reconstructs a segmentation map for intermediate frames. On the YouTube-VOS dataset, our model achieved a J of 63.83 and an inference time of 13.61 msec (73.52 fps). This frame rate is significantly higher than the existing VOS studies in the literature \cite{wang2021survey} while the performance is comparable. 

Our method can also be easily fused with other SOTA VOS studies to increase their frame rate. We integrate our model with RAVOS \cite{miao2022region}; a powerful semi-automatic VOS technique that relies on a previous frame annotation map for future frames. We provide our model's reconstruction as an input annotation frame to RAVOS (for every other frame) and evaluate the overall performance of the RAVOS output. Our experiments show that the new RAVOS model achieves a frame rate of 22.06 fps (9 fps increase from the original 13 fps on GPU II) while the J index only decreases from 79.4 to 74.8 (unseen) and  82.6 to 77.4 (seen) on YouTube-VOS 2019. This 69\% jump in frame rate can provide in applications where a high-speed VOS is needed
\subsection{Other potential applications}
As discussed, a high frame rate depth map can facilitate different tracking tasks, including object tracking. Kowdle \etal \cite{kowdle2018need4} demonstrates several applications of real-time depth estimation, including face tracking, rigid fusion, and hand tracking. Fast object tracking can be applied to many research areas, including healthcare applications such as telesurgery, robotics, human-computer interaction, autonomous driving, \etc. Additionally, one of the main components of efficient real-world perception in VR is depth information. Fast depth sensing is crucial in reducing the lag between what is happening in the real world and the information conveyed to the virtual environment.
High latency results in difficulties in coordinating the interactions between different users, such as passing, tossing and catching objects, shaking hands, or performing tasks with hand \cite{kowdle2018need4}. Utilizing a high frame rate depth estimation technique can dramatically reduce these effects and provide a more realistic experience to VR users. Using the proposed framework in \cref{track}, we built a 3D model in Blender \cite{blender} which can ultimately be used for VR applications.
% \cref{fig:bl} shows the 3D VR model that is being tracked along with the corresponding RGB frame on the right.
% \begin{figure*}
% \centering 
% % \begin{subfigure}{0.68\linewidth} %
% % \fbox{\rule{0pt}{2in}
% \begin{subfigure}{0.49\linewidth}
%     \centering 
%     \includegraphics[width=0.6\linewidth]{\string"bl.png"}
%     \caption{3D model in Blender app}
% \end{subfigure}
% \begin{subfigure}{0.49\linewidth}
%     \centering 
%     \includegraphics[width=.75\linewidth]{\string"blrg.png"}
%     \caption{corresponding RGB frame}
% \end{subfigure}

% % \caption{An example of a subfigure.}
% % \label{fig:short-a} \end{subfigure} \hfill{}\begin{subfigure}{0.28\linewidth}
% % \fbox{\rule{0pt}{2in} \rule{0.9\linewidth}{0pt}} \caption{Another example of a subfigure.}
% % \label{fig:short-b} 
% % \end{subfigure} 
% \caption{Integrating the 3D poses from MediaPipe and AutoDepthNet for tracking VR model in Blender app}
% \label{fig:bl} 
% \end{figure*}

\section{Discussion}
\vspace{-0.2cm}
The proposed method has several characteristics which make it a viable addition to the existing depth cameras.
First, once we have the calibrated RGB-D camera setup, the rest of the pipeline is end-to-end and easy to use. Moreover, considering the fact that high-speed cameras are relatively  common nowadays, a depth camera can easily be coupled with a high-speed RGB camera to make the system work -- future depth cameras could even integrate such a high-speed RGB camera directly. In addition, even though the proposed method is designed for depth reconstruction, the model's design could be used to refine other types of frames; for example, interpolating the result of a slow segmentation model to produce faster image segmentations. 

Colour frames are typically delivered with much lower latency than depth frames, primarily due to the longer capture and processing times involved in depth sensing. The results presented in \cref{past} suggest that we can extrapolate depth frames from two frames in the past without significantly impacting accuracy, thus enabling a latency reduction of up to 59 ms. This has the potential to reduce lag in interactive applications, ranging from video games utilizing body tracking (\eg dance games, exergames) to gestural interfaces in augmented or virtual reality.

\subsection{Inpainting of invalid depth values}
\label{inpaint}
The reconstructed depth map in \cref{fig:pred} demonstrates that the network is able to predict depth for the pixels that are marked as invalid in the ground truth and input image. This feature can be useful as there are many invalidated pixels on the depth camera output, especially around the edges of moving objects which impose limitations for capturing movement between two consecutive frames. However, with the inpainted pixels, we can utilize the depth information at those pixels. \cref{fig:crop} shows a random ROI on a gt image along with the reconstructed image.
Evaluating the accuracy of these pixels is challenging since there is no gt available for these pixels.
%Considering the high precision of the rest of the pixels, one may conclude that the predicted inpainted depth values are also close to the actual depth values as they appear randomly on the image. 
%In order to obtain a thorough evaluation of these pixels, we would need accurate depth values for every single pixel in the scene. 
In future work, we could use a LiDAR sensor to capture every pixel's actual depth with high precision, and thereby evaluate the performance of the system's inpainting abilities.
Another possible way to improve the usability of the inpainted pixels is to use confidence calibration \cite{guo2017calibration} for the network. Using the original invalid pixels mask, we can estimate a confidence level for the reconstructed depth values, which can later be used by the application to determine how best to use the predicted values.

% \begin{figure}[t]
% \centering %
% \begin{subfigure}{0.49\linewidth}
%     \centering 
%     \includegraphics[width=1\linewidth]{\string"inpaintpred.png"}
%     \captionsetup{justification=centering}
%   \caption{reconstructed depth image with inpainted invalid pixels}
%   \label{fig:inp} 
% \end{subfigure}
% \begin{subfigure}{0.49\linewidth}
%     \centering 
%     \includegraphics[width=1\linewidth]{\string"inpaintgt.png"}
%     \captionsetup{justification=centering}
%   \caption{original depth image invalid pixels}
%   \label{fig:inpaint} 
% \end{subfigure}
% \caption{An example of inpainted invalid depth pixels at a random region of interest. The purple pixels on the gt represent invalid pixels (pixels with a depth of 0)}
% \label{fig:cinn} 
% \end{figure}

\subsection{Limitations and future works}
One limitation of the current system is the limited variety of the dataset. Our primary focus was in evaluating the performance of the system for human motion; thus, the dataset consists of a moving person over a static background. For many use-cases, this is sufficient; the model should be capable of copying static depth pixels from the background when the corresponding colour pixels do not change. However, for a more thorough evaluation, and to better highlight the capability of our proposed method, we would ideally expand our dataset significantly to incorporate more complex backgrounds, subjects, and actions.
% While the experimental results highlight the capability of the proposed method on a variety of sequences, we believe that the model needs to be tested with more complicated scenarios where the subject moves at a higher speed. We are also planning on training and testing the model on sequences that contain more complexity and variety in the background, \eg natural scenes, \etc. 

Another shortcoming of the prototype is that it runs offline. This is a limitation of the iPhone: existing streaming tools do not support high-fidelity slow-motion video, so we are limited to retrieving the iPhone's video after recording has completed. In this work, we wanted to focus on the capabilities of current commodity hardware; a future hardware iteration could incorporate a dedicated high-speed camera for real-time operation. In support of real-time use, we reported end-to-end inference times as low as 8 ms, suggesting that high-speed real-time operation is possible.

Finally, although our system does naturally inpaint invalid pixels, we were unable to evaluate the performance of this feature due to missing ground truth data. As noted in the previous section, this could be resolved by using a LiDAR sensor to provide accurate depth data. Anecdotally, inpainted pixels appear to be interpolated from nearby depth values and may not accurately reflect reality; the availability of reliable ground truth for these pixels could also help to train the model for higher inpainting performance.

% Another interesting experiment that we are planning to run is to generate multiple intermediate depth maps between two frames where the depth is available and then use these intermediate maps as input to the network and study how it improves the depth reconstruction.
% Finally, as mentioned in  it is very useful to perform a study for the effectiveness of the network's invalid depth pixel inpainting ass mentioned in \cref{inpaint}
\section{Conclusion}

In this paper, we presented a framework for high frame rate reconstruction of depth maps using commodity sensors. The experimental results suggest that the model is able to effectively reconstruct depth from previous depth frames using the motion information from a high-speed RGB camera. In addition, the proposed method has the ability to inpaint the missing values from the depth camera. We demonstrated the model's utility for body tracking applications, indicating the method's immediate feasibility in domains such as augmented and virtual reality and gaming.

%%%%%%%%% REFERENCES
{\small{} \bibliographystyle{ieee_fullname}
\bibliography{egbib}
 }{\small\par}
\end{document}